\pdfoutput=1

\documentclass[a4paper,fleqn]{cas-sc}

\usepackage[numbers]{natbib}
\usepackage{amsmath}
\usepackage{multirow}
\usepackage{booktabs}
\def\tsc#1{\csdef{#1}{\textsc{\lowercase{#1}}\xspace}}
\tsc{WGM}
\tsc{QE}

\newcommand{\eg}{\emph{e.g.}}

\begin{document}
\let\WriteBookmarks\relax
\def\floatpagepagefraction{1}
\def\textpagefraction{.001}

\shorttitle{Weakly Supervised Scene Text Generation for Low-resource Languages}    

\shortauthors{Yangchen Xie et al.}  

\title [mode = title]{Weakly Supervised Scene Text Generation for Low-resource Languages}  



%

\author[1,5]{Yangchen Xie}
\ead{ycxie0702@gmail.com}
\credit{Conceptualization, Software, Writing - original draft, Writing - review \& editing}

\author[2]{Xinyuan Chen}
\ead{xychen9191@gmail.com}
\credit{Supervision, Validation, Writing - review \& editing}

\author[1,3]{Hongjian Zhan}
\ead{hjzhan@cee.ecnu.edu.cn}
\credit{Supervision, Validation, Writing - review \& editing}

\author[4]{Palaiahankote Shivakumara}
\ead{shiva@um.edu.my}
\credit{Supervision, Validation, Writing - review \& editing}

\author[5]{Bing Yin}
\ead{bingyin@iflytek.com}
\credit{Supervision, Validation, Writing - review \& editing}

\author[5]{Cong Liu}
\ead{congliu@iflytek.com}
\credit{Supervision, Validation, Writing - review \& editing}

\author[1]{Yue Lu}
\ead{ylu@cee.ecnu.edu.cn}
\credit{Supervision, Validation, Writing - review \& editing}
\cormark[1]


\affiliation[a]{organization={School of Communication and Electronic Engineering, East China Normal University},
            city={Shanghai},
            postcode={200241}, 
            country={China}}

\affiliation[b]{organization={AI Laboratory},
            city={Shanghai},
            country={China}}

\affiliation[c]{organization={Chongqing Key Laboratory of Precision Optics, Chongqing Institute of East China Normal University},
            city={Chongqing},
            postcode={401120}, 
            country={China}}

\affiliation[d]{organization={Faculty of Computer Science and Information Technology (FSKTM), University of Malaya},
            city={Kuala Lumpur},
            country={Malaysia}}
            
\affiliation[e]{organization={iFLYTEK Research},
            city={Hefei},
            country={China}}

\cortext[1]{Corresponding author}



\begin{abstract}
A large number of annotated training images is crucial for training successful scene text recognition models. However, collecting sufficient datasets can be a labor-intensive and costly process, particularly for low-resource languages. To address this challenge, auto-generating text data has shown promise in alleviating the problem. Unfortunately, existing scene text generation methods typically rely on a large amount of paired data, which is difficult to obtain for low-resource languages. In this paper, we propose a novel weakly supervised scene text generation method that leverages a few recognition-level labels as weak supervision. The proposed method is able to generate a large amount of scene text images with diverse backgrounds and font styles through cross-language generation. Our method disentangles the content and style features of scene text images, with the former representing textual information and the latter representing characteristics such as font, alignment, and background. To preserve the complete content structure of generated images, we introduce an integrated attention module. Furthermore, to bridge the style gap in the style of different languages, we incorporate a pre-trained font classifier. We evaluate our method using state-of-the-art scene text recognition models. Experiments demonstrate that our generated scene text significantly improves the scene text recognition accuracy and help achieve higher accuracy when complemented with other generative methods.
\end{abstract}


\begin{highlights}
\item A weakly supervised generative method for scene text generation is proposed that utilizes recognition-level annotations for low-resource languages. Also, a cross-language generative scheme is introduced to reduce reliance on labeled data in low-resource languages.
\item We design generative frameworks which employ integrated attention to exploit the global and local relationships between between content features and generated features.
\item With the proposed methods, we generate a large-scale scene text dataset for low-resource languages, which is used to train a scene text recognizer. Our approach significantly improved the performance of the recognizer.
\end{highlights}

\maketitle

\section{Introduction}
\label{introduction}
Text is fundamental to preserving and transferring information in our daily life. Due to the diversity of angles, shapes, and backgrounds, obtaining textual information from scene images has been a challenging task for many years. Recently, with the development of deep learning, the performance of the scene text recognition model has improved remarkably by learning from millions of labeled data \cite{baek2019STRcomparisons,
DBLP:journals/eswa/NaiemiGK21,DBLP:journals/pami/WangXLLLYLS22,DBLP:journals/eswa/XiaoNSC22,  DBLP:journals/eswa/ZhongLSPL22}. However, the majority of the existing scene text dataset is collected for high-resource languages such as English and Chinese. There are very few public scene text data sets for low-resource languages that are also used by a large population of people. As a result, the recognition performance of low-resource languages does not meet the requirements. Furthermore, collecting and annotating scene images of low-resource languages requires experts, which is more expensive than labeling a high-resource language dataset.

Recently, researchers employ synthesis algorithms to alleviate the shortage of labeled scene text images from the real world. MJ\cite{DBLP:journals/corr/JaderbergSVZ14} and ST\cite{DBLP:conf/cvpr/GuptaVZ16}, which are widely used in successful scene text recognition models, are two notable synthesis engines. MJ introduces several rendering modules to generate cropped word images, and each module focuses on features of different parts, such as the font, border/shadow, base color, projective distortion, natural data blending, and noise. ST identifies the suitable regions and then merges the background regions and foreground text via Poisson image editing. Also, foreground text is rendered by font, border/shadow, and color. These methods apply the existing computer fonts and corpus to generate foreground text, which is then merged with diverse backgrounds using image editing to generate infinite amounts of paired data. However, designing computer fonts requires human experts and there are few computer fonts for low-resource languages. For instance, there are 1394 Latin fonts in Goole Fonts, but only 197 Cyrillic and 31 Korean fonts. In addition, Peng Z et.\cite{DBLP:conf/cvpr/ZhouHMD18} assumed that simply merging text and background in traditional image editing methods cannot simulate the distribution of real data. After that, \cite{DBLP:journals/chinaf/LiaoSLHYB20, DBLP:journals/corr/abs-2003-10608, DBLP:conf/icdar/YimKCP21} improve the performance of the synthesis engines and extend to more optimized and more diverse scenarios, but they still rely heavily on computer fonts and can not be applied to low-resource languages.

\begin{figure*}[t]
\label{fig1}
	\centering
    \includegraphics[width=1\linewidth]{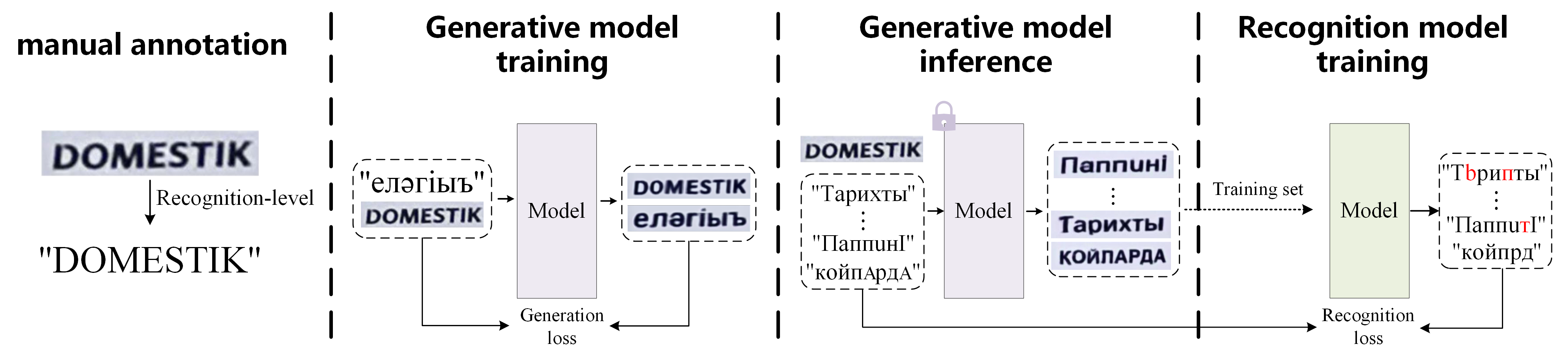}
	  \caption{\textbf{The pipeline of training scene text recognition models based on the proposed methods}. (a) Images are annotated at the character level. (b) the generative model is trained using annotated datasets from previous steps. (c) The trained model replaces text in scene text images from high-resource languages by leveraging the provided text from low-resource languages. (d) The recognition model is trained using synthetic datasets.}
\label{fig:1}
\end{figure*}

Some other researchers exploit scene text editing methods to generate scene text.  \cite{DBLP:conf/mm/WuZLHLDB19, DBLP:conf/cvpr/YangHL20} introduce two conversion modules to process foreground text and background images separately. These methods apply GAN\cite{DBLP:conf/nips/GoodfellowPMXWOCB14} to fuse the processed text and text-erased images such that the distribution of generated images can converge to the true distribution. Due to the requirement for target-style images, these methods are limited to training on synthetic data. Rewrite\cite{DBLP:journals/corr/abs-2107-11041} and TextStyleBrush\cite{DBLP:journals/corr/abs-2106-08385} introduce a pre-trained OCR to ensure the correctness of the generated text and employ real-world images for training. With real-world images, models can imitate more text styles from scene images. However, these methods are restricted to high-resource languages such as English and Chinese. A robust and accurate OCR is hard to obtain for low-resource languages. Inspire by image-to-image translation methods,  
 DGFont\cite{DBLP:conf/cvpr/XieCSL21}  and DGFont++\cite{DBLP:journals/corr/abs-2212-14742} introduce a novel deformable module that deforms and transforms the character of one font to another in an unsupervised way. These methods are promising in addressing generation problems for low-resource languages. However, on the one hand, DG-Font is designed for a "character" in black and white, while scene text images often contain "text" with diverse backgrounds. On the other hand, DG-Font can only be applied to specific languages, which cannot fully leverage the existing scene text images from high-resource languages. 

Compelled by the above observations, we propose a novel weakly supervised scene text generation method. The main idea of the proposed method is to train scene text generation models based on a few recognition-level annotations for low-resource languages. As illustrated in Figure \ref{fig1}, the recognition-level labels are weak supervision; the text of the images are known but the target edited images are unknown. With the recognition-level labels of the given content and scene text images as weak supervision, the proposed method learns to separates the content and style features respectively and then mixes the two domain representations to generate target scene text images. The content is defined as textual information from images and the style is defined as scene text characteristics such as font, alignment, and background. We introduce an integrated attention module to process features from the content encoder at different levels respectively. For low-level features, a sequence of densely connected deformable convolutions is applied to perform global attention. We first employ the 'query' features from the content encoder and 'key' features from the decoder to predict pairs of displacement maps. The predicted maps are then applied to these dense deformable convolutions to process the 'value' features from the content encoder. For high-level features, a deformable convolution followed by local attention is introduced to predict the local relationship between the encoder and mixer features. Moreover, to further leverage the existing datasets of high-resource languages, we add data from high-resource languages into training and perform inference in a cross-languages way. Specifically, in the training process, images of high-resource languages help train the scene text generation model by increasing the training set. In the testing process, the model replaces text in scene text images from high-resource languages by leveraging the provided text from low-resource languages, which helps generate datasets of diverse scenes. Besides, a pre-trained font classifier is introduced to maintain the consistency of style in different languages. Extensive experiments demonstrate that our model outperforms the state-of-the-art scene text generation methods in improving scene text recognition accuracy. Moreover, results show that our method is able to be complementary to commonly used generation methods.

In summary, our contributions can be summarized as follows:
\begin{enumerate}[(a)]
\item A weakly supervised generative method for scene text generation is proposed that utilizes recognition-level annotations for low-resource languages. Also, a cross-language generative scheme is introduced to reduce reliance on labeled data in low-resource languages.
\item We design generative frameworks which employ integrated attention to exploit the global and local relationships between between content features and generated features.
\item With the proposed methods, we generate a large-scale scene text dataset for low-resource languages, which is used to train a scene text recognizer. Our approach significantly improved the performance of the recognizer.
\end{enumerate}  

\section{Related works}
\subsection{Scene Text Recognition}
Automatic Scene text recognition of various texts in scene images has been a popular research topic and an important task in a wide range of industrial applications for many years \cite{DBLP:journals/csur/ChenJZLW21,DBLP:journals/pami/YeD15}. Early researches focus on hand-crafted features\cite{DBLP:conf/cvpr/NeumannM12, DBLP:journals/tip/YaoBL14}. With the development of deep learning, the scene text recognition models have been well explored and achieved great success in the past few years \cite{baek2019STRcomparisons, baek2019STRcomparisons, DBLP:journals/eswa/NaiemiGK21, DBLP:journals/eswa/XiaoNSC22, DBLP:journals/eswa/ZhongLSPL22}. However, training scene text recognition models requires a large number of labeled images which is extremely difficult to collect for low-resource languages. The accuracy of English scene text recognition is over 90 percent \cite{baek2019STRcomparisons} in the regular datasets. In contrast, there are few reports on low-resource languages like Kazakh. On the one hand, existing low-resource languages data sets such as ICDAR2017 \cite{DBLP:conf/icdar/GomezSGNVMBK17} and ICDAR2019 \cite{DBLP:conf/icdar/NayefLOPBCKKM0B19} contain only a few hundred or thousand training samples which are far from being able to train scene recognition models. On the other hand, collecting and annotating scene images of low-resources languages requires experts, which is expensive. In this work, the proposed universal scene text generation method tries to address this challenge by generating a large amount of realistic scene text images. The proposed generation methods is evaluated based on the state-of-the-art recognition model \cite{baek2019STRcomparisons}.

\subsection{Data augmentation}
Data augmentation is an important technique to avoid overfitting when training deep neural networks, especially in the shortage of training data. Typical image augmentations include two kinds of transforms, spatial levels such as cropping, rotation, perspective, and resizing, and pixels such as noise, blur, and color change. Existing scene text recognition methods often select a subset of these augmentations in the training process. Recently, some researchers explore data augmentations that are suitable for text recognition models. \cite{DBLP:journals/access/MuSXL21} propose Random Blur Region and Random Blur Units to help train more robust deep models. \cite{DBLP:conf/iccvw/Atienza21} analyze the difference in data augmentation between object recognition and text recognition and formulates a library of data augmentation function to improve the performance of existing recognition methods. However, data augmentation cannot solve the data shortage problem of low-resource languages. The aforementioned data augmentation strategy is often based on experiments of English datasets. Identification of optimal augmentations for low-resources is a challenge. Besides, the robust scene text recognition methods rely on lots of scene text images with diverse backgrounds and fonts, which cannot be achieved by using data augmentation.

%

\subsection{Scene Text Generation}
Scene text generation aims to automatically generate scene text images based on given textual content. SynthText \cite{DBLP:conf/cvpr/GuptaVZ16} is a notable scene text image generator and has been commonly applied to train scene text recognition models. It first analyzes images with off-the-shelf models and searches for suitable text regions on semantically consistent regions to put processed text from the specific font. After that, SynthText3D \cite{DBLP:journals/chinaf/LiaoSLHYB20} and its improved version UnrealText \cite{DBLP:journals/corr/abs-2003-10608} synthesize realistic scene text images from the unreal world via a 3D graphics engine. However, these methods rely heavily on computer fonts designed by human experts. There are few computer fonts for Low-resource languages. Some image composition methods \cite{DBLP:conf/cvpr/ZhanZL19,DBLP:journals/corr/abs-1901-09193} introduce generative adversarial networks to make synthesized images more realistic, but they still suffer from the reliance of computer fonts. 

Recently, there are some attempts to generate scene text by scene text editing methods. Initial works \cite{DBLP:conf/mm/WuZLHLDB19,DBLP:conf/cvpr/YangHL20} train the model to separate the text region and the background region using the text-erased image. These methods require target-style images for training, thereby being restricted to training on synthetic data. STEFANN \cite{DBLP:conf/cvpr/RoyBG020} proposed a font adaptive neural network that replaces individual characters in the source image using a target alphabet. This approach assumes per-character segmentation which is impractical in many real-world images. Rewrite \cite{DBLP:journals/corr/abs-2107-11041} and TextStyleBrush \cite{DBLP:journals/corr/abs-2106-08385} introduce self-supervised training schemes and employ real-world images in the training process. However, these methods rely on the pre-trained OCR models to constrain the content of the generated images, and low-resource languages often lack enough images to train robust OCR models.

\subsection{Unsupervised Image-to-image 
translation}
The purpose of image-to-image translation is to learn the mapping from the image in the source domain to the target domain, and it has been applied to many image generation fields, such as artistic style transfer \cite{DBLP:conf/eccv/JohnsonAF16,DBLP:conf/eccv/ZhangD18}, video frame generation \cite{DBLP:conf/iccv/ChanGZE19,DBLP:journals/tip/ChenXYT20}, and font generation \cite{DBLP:conf/cvpr/XieCSL21}. Pix2pix \cite{DBLP:conf/cvpr/IsolaZZE17} is the first image-to-image conversion model based on the conditional GAN. To solve the unpaired image-to-image conversion, FUNIT \cite{DBLP:conf/iccv/0001HMKALK19} and TUNIT \cite{DBLP:journals/corr/abs-2006-06500} assume that animal images can be decomposed into content and style features, and these features can be recombined through adaptive instance normalization \cite{DBLP:conf/iccv/HuangB17}. DG-Font \cite{DBLP:conf/cvpr/XieCSL21} further proposes an unsupervised font generation method that does not require human experts and can be applied to any language system. It proved that using deformable convolution as the skip connection helps maintain the integrity of the content in generated images. Our work is related to image-to-image translation when we assume that the scene text image can be decomposed into content and style features. However, the above-mentioned image-to-image conversion method usually defines a fixed number of styles and cannot be directly applied to scene text generation. Besides, a carefully designed spatial attention mechanism is proved to improve the quality of generated images \cite{DBLP:conf/cvpr/YinSL21,DBLP:conf/icml/ZhangGMO19}. Inspired by the previous work, we propose a weakly supervised scene text generation method.


\begin{figure*}[t]
\begin{center}
   \includegraphics[width=\linewidth]{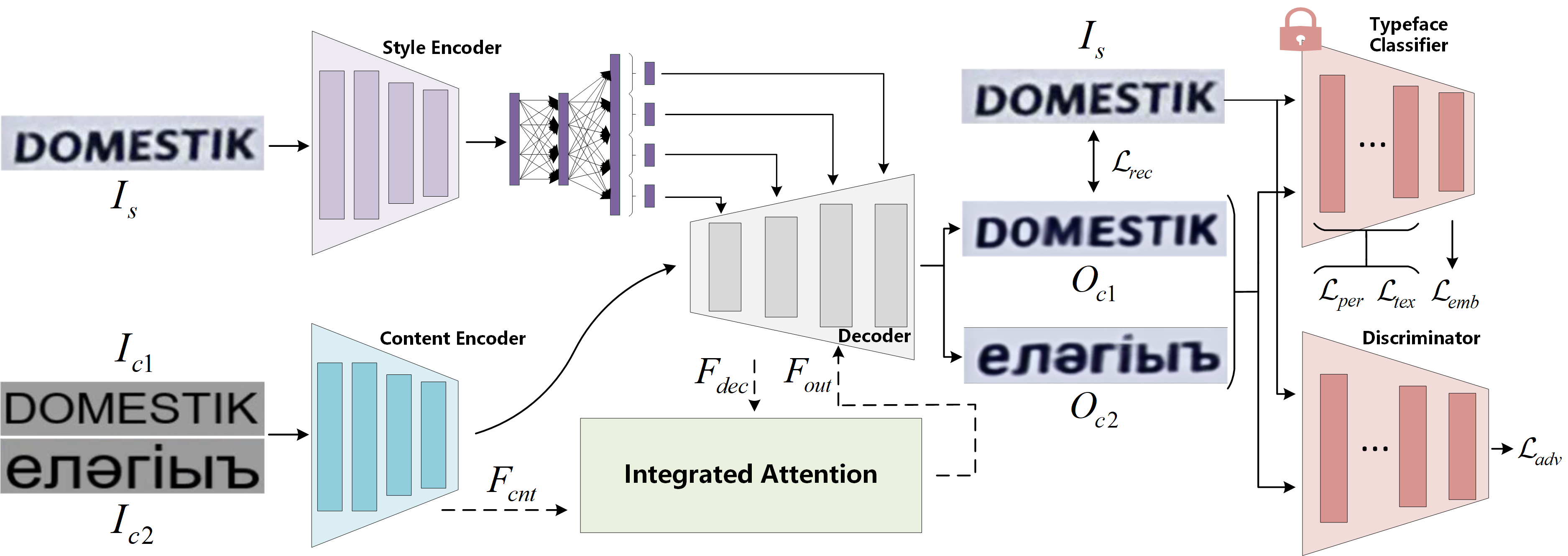}
\end{center}
   \caption{\textbf{Overview of the proposed method.} The style encoder and content encoder encode the scene text images and content images respectively. The integrated attention module takes the content features $F_{cnt}$ and decoder features $F_{dec}$ as input and output the $F_{out}$ which is then concatenated with $F_{dec}$. The output images are input to the typeface classifier and discriminator to distinguish the style and reality.}
   \label{fig2}
\end{figure*}

\section{Methods}
\subsection{Overview}
Given the content images $I_c$ and scene text images $I_s$, the proposed model aims to replace the text in the $I_s$ with the given content string while maintaining the original style. We propose a novel scene text generation method by utilizing recognition-level labels as weak supervision. As shown in Figure \ref{fig2}, the generative model consists of a content encoder, a style encoder, a decoder, and an integrated attention module. To this end, we use a content encoder and a style encoder: for content, ($E_{c}$), and style, ($E_{s}$). Thus, content representation is produced by $F_{cnt} = E_c(I_c)$ and style representation by $F_{sty} = E_s(I_s)$. Given the style images $I_s$ cropped from scene images, we employ the recognition-level labels as weak supervision and render it using a standard font for the text, on a plain grey background, producing the image $I_{c1} \in R^{H \times W}$. Besides, another image $I_{c2}$ with random text is produced in the same way. The proposed method extracts latent style and content representations based on $I_s$ and $I_c$. The style vectors are mapped to AdaIN normalization coefficient \cite{DBLP:conf/iccv/HuangB17} through several fully connected layers. The generator is designed to generate the edited scene text image features $F_{dec}$ by mixing these two representations. Besides, to generate images with complete content structure, the integrated attention module aims to learn the global and local relations of features between content features and generated features from different layers. Details are described in Sec \ref{3.2}. When the images are generated from the generative networks, a typeface classifier pre-trained on a set of synthetic fonts and a discriminator are introduced to distinguish the style and reality between scene text images $I_s$ and generated images $O_{c1}$, $O_{c2}$.

\subsection{Integrated Spatial Attention Module}
\label{3.2}
\begin{figure*}[t]
\begin{center}
   \includegraphics[width=\linewidth]{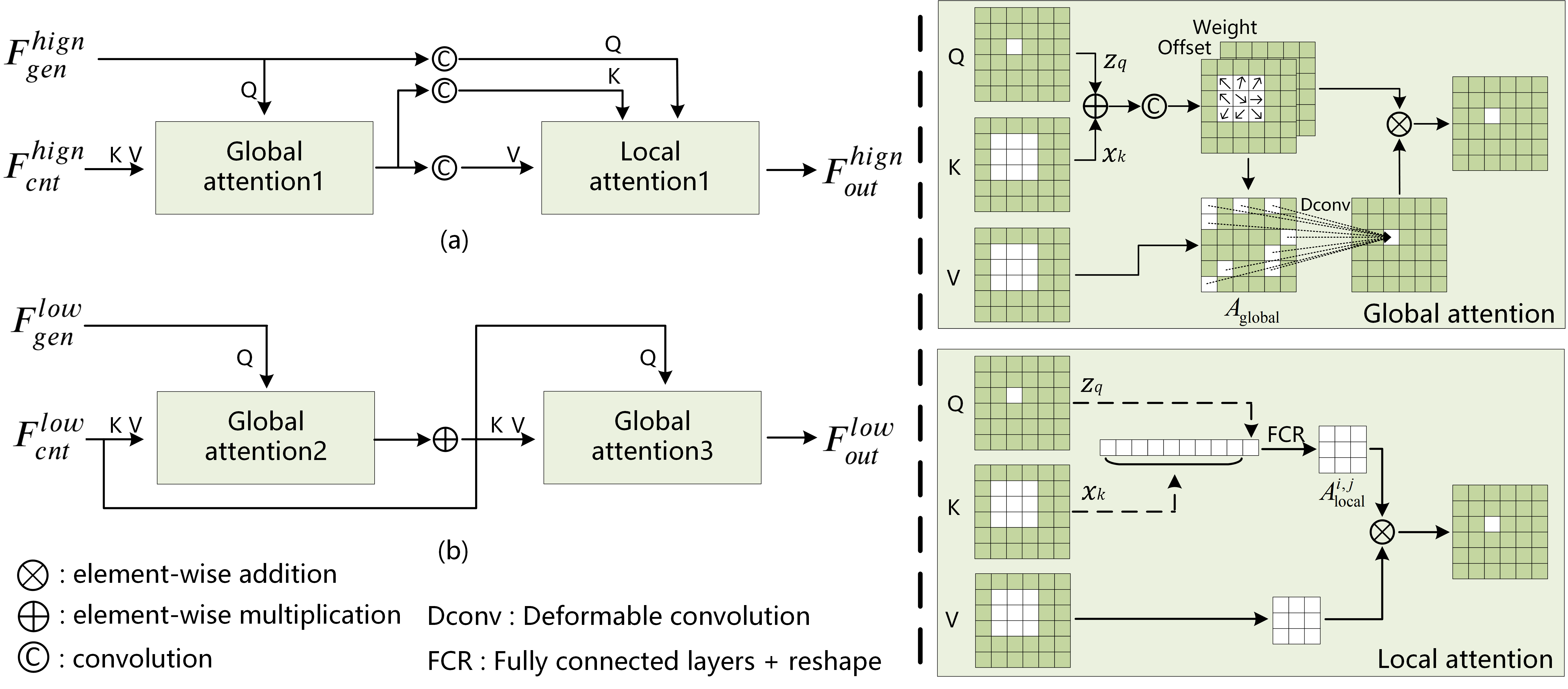}
\end{center}
   \caption{\textbf{Details of the integrated attention.}}
   \label{fig3}
\end{figure*}

The proposed weakly supervised approach reduces the annotation cost of scene text generation for low-resource languages. But the content of generated images is prone to missing some parts. Compelled by the observation that there exists deformation and stroke mapping between content image and generated image, we design an integrated attention module to ensure the complementation of generated content at both global and local levels. Specifically, the global attention which is modeled by deformable convolution learns the point-wise deformation and deforms the source content features by learning a global sparse weight $A_\text{{global}}$. The local attention learns the local stroke mapping between the source content feature and the target generator feature by learning a local dense attention weight $A_\text{{local}}$. As shown in figure \ref{fig3}, the content features $F_{cnt}$ from different levels are processed in different ways. In figure \ref{fig3} (a), for the high-level features, the global attention is first adopted to help deform the content feature $F_{cnt}^{high}$ to $F_{dec}^{high}$. After that, a local attention module is employed to learn the local spatial mapping between content images and generated images, such as strokes and radical decomposition, which exist in high-level features. Different from the high-level features, low-level features are often minor details of the images, like lines or dots, and have similar spatial properties as the original input images. For the low-level features in Figure \ref{fig3}(b), we adopt a densely connected global attention module to deform the features $F_{cnt}^{low}$. 

Attention module measure the similarity of query-key pairs by the compatibility function of the query with a set of key elements and can be viewed from a unified perspective  \cite{DBLP:conf/iccv/ZhuCZLD19}. Specifically, $q$ indicates the index of a query element with the content $z_q$, and $k$ indicates the index of a key element with the content $x_k$. The output of the attention module is computed as:
\begin{equation}
\small
y_q = W^{'}\sum_{k\in \Omega_q}A(q,k,z_q,x_k) \odot Wx_{k},
\label{eq:general_attention}
\end{equation}

where $\Omega_q$ indicates the supporting key region for the query, $A(q,k,z_q,x_k)$ indicates the attention weights or the similarity, and $W^{'}$ and $W$ indicate the learnable weights. The output $y_q$ is calculated by element-wise multiplication of $W^{'}$ multiplied by the attention weights and $Wx_k$. Besides, the similarity are usually normalized based on $\Omega_q$m like $\sum_{k\in \Omega_q}A(q,k,z_q,x_k) = 1$. In our model, $q_k$, $x_k$, and $Wx_k$ denote the sampled features from query $Q$, key $K$, and value $V$ respectively, and $W^{'}$ denotes a convolution or the control weight of the output.

The global attention employs the learnable offsets to adjust the sampling positions of key elements to capture the global spatial relationship. These target key elements can be sampled from any sampling position of the input feature map. The learnable offsets are predicted based on the query content $q_k$ and key content $x_k$ and are dynamic to the input. When incorporating the global attention into the Eq. \ref{eq:general_attention}, the attention weights is:
\begin{equation}
\small
A_{\text{global}}(q,k,z_{q},x_{k}) = interp(k, q+p+w^{\top} \cdot concat(z_{q}, x_{k})),
\end{equation}
where $q$, $k$ are the index of feature map, $p$ are the sampling offset in regular convolution. $w^{\top}$ is modeled by a regular convolution layer and projects the concatenation query content $z_q$ and the key content $x_k$ to a learned offset. Follow \cite{DBLP:conf/cvpr/ZhuHLD19}, $interp$ is the bilinear interpolation kernel. For the global attention, the weight $W$ is fixed as identity and $W^{'}$ is a convolution layer that samples on the irregular and offset locations.

For local attention, the module is different from traditional transformer attention in \cite{DBLP:conf/nips/VaswaniSPUJGKP17} and predicts the weight of a location relative to the features of its neighbors, rather than the entire input feature. Specifically, for each position $(i, j)$ in spatial dimensions of the query element $z_q$, denoted as $z_q^{i,j} \in  \mathbb{R}^{1 \times 1 \times c}$, we extract a patch with size $s$ centered at $(i, j)$ from the corresponding key element$x_k$, denoted as $x_k^{i,j} \in  \mathbb{R}^{s \times s \times c}$. Then the weight $A^{i,j}_{\text{local}(q,k,z_q,x_k)} \in \mathbb{R}^{s \times s \times c}$ is obtained by reshaping the estimation of a fully connected network(FCN):
\begin{equation}
    A^{i,j}_{\text{local}(q,k,z_{q},x_{k})}=reshape(FCN(concat(flatten(x_{k}^{i,j}),z_{q}^{i,j}))).
\end{equation}
The flattened patch query $flatten(k)$ and the corresponding query $z_{q}^{i,j}$ are first concatenated, and compute the FFN function. The FFN function is composed of a fully-connected layer followed by leaky ReLu and a linear fully-connected layer. We iterate over all the $(i,j)$ to constitute the output of the local spatial attention module. For the local attention, $W$ represents the mathcal transformation between $Q$ and $K$, and $W^{'}$ is the control weight of the output.


\subsection{Typeface Classifier}
We leverage existing datasets from high-resource languages in both training and testing processes to help synthesize datasets with diverse backgrounds and font styles. In order to help the generator captures the style of input text regardless of its language, we introduce a pre-trained typeface classifier inspired by \cite{DBLP:journals/corr/abs-2106-08385}. Specifically, 
the classifier is trained on synthetic datasets of high-resource languages, which is generated by \cite{DBLP:journals/corr/JaderbergSVZ14}. The classifier is trained to identify the synthetic fonts and is only used to provide a perceptual signal for training the generative models. 

The detail of the synthetic data used to train the classifier is described in Sec \ref{4.2}. Given a word image, a VGG19 network is employed to classify its style. We first encode font class into one-hot labels and train the networks with softmax loss. After the classifier is trained, we employ the network as supervision and compute the style alignment loss $\mathcal{L}_{type}$. Specifically, given a word image, the loss is as follow:
\begin{gather}
\mathcal{L}_{type} = \lambda_{1}\ell_{per} + \lambda_{2}\ell_{tex} + \lambda_{3}\ell_{emb},\\
\ell_{per}=\mathbb{E}[\sum_i\frac{1}{M_i}\parallel \phi_i(I_{s})-\phi_i(O_{c})\parallel_1],\\
\ell_{tex}=\mathbb{E}_i[\parallel G_i^\phi(I_{s}) -G_i^\phi(O_{c})\parallel_1],\\
\ell_{emb}=\mathbb{E}[\parallel \psi(\mathcal{I}_{s})-\psi(\mathcal{O}_{c})\parallel_1].
\end{gather}
The perceptual loss $\ell_{per}$ is computed from the feature maps at layer i denoted as $\phi_i$. We normalize the loss using the number of elements in the particular feature map, denoted by $M_i$. We also compute a texture loss, $\ell_{tex}$, from the Gram matrices $G_i^{\phi}=\phi_i\phi_i^T$ of the feature maps. These losses capture the background style information and are computed on the output features of relu1$\_$1, relu2$\_$1, relu3$\_$1, relu4$\_$1, and relu5$\_$1. Besides, the embedding loss $\ell_{emb}$ is computed on the feature maps of penultimate layers $\psi$ and mainly learns the font style of the generated image.

\subsection{Loss Function}
Our model aims to achieve automatic scene text generation via a weakly supervised method. We adopt four losses: 1) adversarial loss is used to produce realistic images; 2) content consistent loss is introduced to encourage the content of the generated image to be consistent with the content image; 3) image reconstruction loss is used to maintain the domain-invariant features; 4) style alignment loss to bridge the style gap between generate images and scene text images. We describe the formula of these losses and the full objective in this section.

\textbf{Adversarial loss: } our model aims to generate plausible images by solving a mini-max optimization problem. The generative network $G$ tries to fool discriminator $D$ by generating fake images. The adversarial loss penalty the wrong judgment when real/generated images are input to the discriminator. The hinge GAN loss\cite{DBLP:conf/icml/ZhangGMO19} was used:
\begin{gather}
\mathcal{L}^D_{adv} = -\mathbb{E}_{I_s, y \in P_s, I_c \in P_c} \text{min}(0, -1+D (O_c, y))  - \mathbb{E}_{I_s, y \in P_s, I_c \in P_c} \text{min}(0, -1-D(G(I_s, I_c), y)),\\
\mathcal{L}^G_{adv} = - \mathbb{E}_{I_s, y \in P_s, I_c \in P_c} D(G(I_s, I_c), y).
\end{gather}

\textbf{Content consistent loss: }adversarial loss is adopted to help the model to generate a realistic style while ignoring the correctness of the content. To prevent mode collapse and ensure that the features extracted from the same content can be content consistent after the content encoder $f_c$, we impose a content consistent loss here:
\begin{equation}
\label{Lcnt}
\mathcal{L}_{cnt} =  \mathbb{E}_{I_s \in P_s, I_c \in P_c}\left\|Z_c - f_c(G(I_s, I_c))\right\|_1 .
\end{equation}
${L}_{cnt}$ ensures that given a source content image $I_c$ and corresponding generated images, their feature maps are consistent after the content encoder $f_c$.

\textbf{Image reconstruction loss: }adversarial loss is adopted to help the model generate realistic styles while focusing on high-frequency. With recognition-level labels,  the generator can reconstruct the source image $I_c$ when given its origin style, we impose a reconstruction loss:

\begin{equation}
\label{Lrec}
\mathcal{L}_{rec} = \mathbb{E}_{I_c \in P_c}\left\|I_c - G(I_{s},O_{c})\right\|_1 .
\end{equation}

The objective helps preserve domain-invariant characteristics (\eg content) of its input image $I_c$.

\textbf{Full objective loss: } combining all the loss, our full objective loss is given as follow:
\begin{equation}
\label{Limg}
\mathcal{L} = \max_{D}\min_{G}\mathcal{L}_{adv} + \lambda_{img}\mathcal{L}_{img} + \lambda_{cnt}\mathcal{L}_{cnt} + \lambda_{sty1}\mathcal{L}_{sty}(O_1) + \lambda_{sty2}\mathcal{L}_{sty}(O_2),
\end{equation}  
where $\lambda_{img}$, $\lambda_{cnt}$, $\lambda_{sty1}$, and $\lambda_{sty2}$ are parameters to control the weight of each objective. The details of these hyper-parameters are report in Sec \ref{4.1}.

\section{Experiments}
In this section, we evaluate our proposed model based on scene text recognition tasks for Korean and Kazakh. The implementation details are described first. We then introduce our dataset. After that, the results of our experiments are shown to verify the advantages of our model.

\begin{table*}[ht]
\caption{\textbf{Architecture of Generative networks.}}
\begin{center}
\begin{tabular}{cccccccc}
\hline
\hline
                                 & Type                              & Kernel size & Resample & Padding & Feature maps & Normalization & Nonlinearity \\ \hline
\multirow{8}{*}{Style encoder}   & Convolution                            & 3           & MaxPool  & 1       & 64           & BN            & ReLU         \\
                                 & Convolution                            & 3           & MaxPool  & 1       & 128          & BN            & ReLU         \\
                                 & Convolution                            & 3           & - & 1       & 256          & BN            & ReLU         \\
                                 & Convolution                            & 3           & MaxPool  & 1       & 256          & BN            & ReLU         \\
                                 & Convolution                            & 3           & - & 1       & 512          & BN            & ReLU         \\
                                 & Convolution                            & 3           & MaxPool  & 1       & 512          & BN            & ReLU         \\
                                 & Convolution                            & 3           & - & 1       & 512          & BN            & ReLU         \\
                                 & Convolution                            & 3           & MaxPool  & 1       & 512          & BN            & ReLU         \\ 
                                 & Average pooling                                     & -           & -        & -       & 128          & -             &  -   \\
                                 & FC                                     & -           & -        & -       & 128          & -             &  -   \\
                                 \hline
                                 
\multirow{4}{*}{Content encoder} & Deform. conv.                        & 7           & - & 3       & 64           & IN            & ReLU         \\
                                 & Deform. conv.                        & 4           & stride-2 & 1       & 128          & IN            & ReLU         \\
                                 & Deform. conv.                        & 4           & stride-2 & 1       & 256          & IN            & ReLU         \\
                                 & Res block $\times$ 2 & 3           & - & 1       & 256          & IN            & ReLU         \\ \hline
                                 
\multirow{4}{*}{Decoder}           & Res block $\times$ 2 & 3           & - & 1       & 256          & AdaIN         & ReLU         \\
                                 & Convolution                            & 5           & Upsample & 2       & 128          & AdaIN         & ReLU         \\
                                 & Convolution                            & 5           & Upsample & 2       & 64           & AdaIN         & ReLU         \\
                                 & Convolution                            & 7           & - & 3       & 3            & -             & tanh       \\ \hline
\end{tabular}
\end{center}
\label{architecture1}
\end{table*}

\subsection{Implementation Details}
\label{4.1}
the weights of convolution layers are initialized with He initialization \cite{DBLP:conf/iccv/HeZRS15}, whereby the biases are set to zero, and the weights of the linear layers are sampled from a Gaussian distribution with mean 0 and standard deviation 0.01. We use the Adam optimizer with $\beta_1=0.9$ and $\beta_2=0.99$ for the style encoder, while the content encoder and decoder are optimized using the RMSprop optimizer with $\alpha$=0.99. The model is trained for 200 epochs with a learning rate of 0.0001 and a weight decay of 0.0001. The hinge adversarial loss \cite{DBLP:conf/icml/ZhangGMO19} is used, with R1 regularization \cite{DBLP:conf/icml/MeschederGN18} using $\gamma=10$. We empirically set the weights of different hyper-parameters as follow: $\lambda_1$ = 1, $\lambda_2$ = 250, $\lambda_3$ = 1, $\lambda_{cnt}$ = 1, $\lambda_{img}$ = 10, $\lambda_{sty1}$ = 1, and $\lambda_{sty1}$ = 0.1.


With regard to the generation model, the batch size is set to 16 and 
we resize the text image height to 64 and keep the same aspect ratio. In the training process, we randomly select the batch data and resize these images to the average width, and during testing, we input images of variable width to get the desired results. In the integrated attention, $F_{cnt}^{high}$ and $F_{gen}^{high}$ are extracted from the second downsampling and penultimate upsampling layer, and $F_{cnt}^{low}$ and $F_{gen}^{low}$ are extracted from first downsampling and the last upsampling layer. Table \ref{architecture1} shows the details of our generative networks, including both the encoder components and decoder components. BN, IN, AdaIN denote batch normalization, instance normalization, and adaptive instance normalization, respectively. FC means the fully connected layers. To evaluate our method, we employ three recognition methods. The batch size is 256 and none of these methods are trained with data augmentation unless otherwise specified.

\subsection{Dataset and Evaluation Metrics}
\label{4.2}
To evaluate our model for scene text recognition in low-resource languages, we choose Kazakh (84 characters) and Korean (2180 characters) to represent the low-resource languages in alphabetical language and logographic language respectively and English and Chinese to represent the high-resource languages to help train our model. For the cases of Kazakh, we collect a dataset that contains 22,182 Kazakh images and 81,900 English images for training, and 4571 Kazakh images for testing. For the cases of Korean, the training set consists of a total of 16,279 Korean images and 113,491 Chinese images, and the test set contains 4644 Korean images cropped from ICDAR2019-MLT\cite{DBLP:conf/icdar/NayefLOPBCKKM0B19}. In our experiments, all the training and testing images are box images cropped from real scene images. Besides, the training sets for typeface classifiers are generated by SynthText, which employs 284 Chinese fonts and 800 English fonts, respectively.  

In this paper, we evaluate the effectiveness of synthetic data in training scene recognition models. To this end, we generate datasets using different methods and evaluate the trained models on real scene text images.  For each experiment, a total of 1 $M$ box images generated by one of the comparison methods are used for training the recognition models. In this study, we investigate the contribution of generated images in training recognition models. As a result, we report the accuracy and normalized edit distance \cite{DBLP:conf/icdar/ShiYLYXCBLB17} on the testing set instead of FID score \cite{DBLP:conf/nips/HeuselRUNH17} that are commonly used in image generation task. Specifically, the normalized edit distance is computed as:
\begin{equation}
\label{edit_distance}
Norm =  1-D(s_i,\hat{s_i}),
\end{equation}
where $D(:)$ indicates the Levenshtein Distance. $s_i$ and $\hat{s_i}$ indicate the predicted text line in string and the corresponding ground truths. 

\begin{table*}[t]
\caption{Compared with the state-of-the-art Methods based on TRBA\cite{baek2019STRcomparisons}}.
\begin{tabular}{lllll}
\toprule
Training data & Kazakh(Acc) & Kazakh(ED) & Korean(Acc) & Korean(ED) \\
\midrule
Real dataset  & 54.295 & 0.799 & 40.909& 0.662 \\
Real dataset(aug)\cite{DBLP:journals/pami/WangXLLLYLS22}  & 59.899 & 0.830 & 48.971 & 0.714 \\
\midrule
SRNet\cite{DBLP:conf/mm/WuZLHLDB19}  & 34.697 & 0.644 & - & - \\
TUNIT\cite{DBLP:journals/corr/abs-2006-06500} & 55.874 & 0.800 & 17.539 & 0.444\\
DGFont\cite{DBLP:conf/cvpr/XieCSL21} & 60.053 & 0.820 & 39.237 & 0.625 \\
Ours  & 71.319 & 0.873 & 64.837 & 0.806 \\
\midrule
SynthText\cite{DBLP:conf/cvpr/GuptaVZ16} & 62.248 & 0.799 & 68.418 & 0.820 \\ 
SynthText+SRNet   & 43.251  & 0.711  & -  & -  \\
SynthText+TUNIT   & 68.366  & 0.853 & 66.467 & 0.818  \\
SynthText+DGFont  & 70.750  & 0.859  & 69.554 & 0.834  \\
SynthText+Ours    & 73.047  & 0.877 & 72.706 & 0.849 \\
\bottomrule
\end{tabular}
\label{quantatitive_results}
\end{table*}

\begin{table*}[t]
\caption{Evaluate the generalizability of the proposed methods}
\label{Recognition_table}
\begin{tabular}{llllll}
  \toprule 
  Method & Training data & Kazakh(Acc) & Kazakh(ED) & Korean(Acc) & Korean(ED) \\
  \midrule
  \multirow{4}{*}{TRBA \cite{baek2019STRcomparisons}} & Real dataset  & 54.295 & 0.799 & 40.909 & 0.662\\ 
   & Ours & 71.319 & 0.873 & 64.837 & 0.806\\ 
   & SynthText & 62.248 & 0.799 & 68.418 & 0.820\\
   & SynthText+Ours & 73.047 & 0.877 & 72.706 & 0.849\\
  \midrule  
  \multirow{4}{*}{PARSeq \cite{DBLP:conf/eccv/BautistaA22}} & Real dataset*  & 56.836 & 0.808 & 37.263 & 0.627\\ 
   & Ours & 71.188 & 0.870 & 64.415 & 0.804\\ 
   & SynthText & 67.928 & 0.849 & 69.754 & 0.832\\
   & SynthText+Ours & 74.491 & 0.883 & 74.874 & 0.871\\
   \midrule  
  \multirow{4}{*}{SCATTER \cite{DBLP:conf/cvpr/LitmanATLMM20}}& Real dataset  & 63.728 & 0.875 & 48.395 & 0.715\\ 
   & Ours & 77.182 & 0.932 & 64.301 & 0.797\\ 
   & SynthText & 72.982 & 0.923 & 66.981 & 0.823\\
   & SynthText+Ours & 80.398 & 0.949 & 73.768 & 0.855\\
  \bottomrule
\end{tabular}
\label{general}
\end{table*}

\begin{figure*}[htbp]
\begin{center}
   \includegraphics[width=\linewidth]{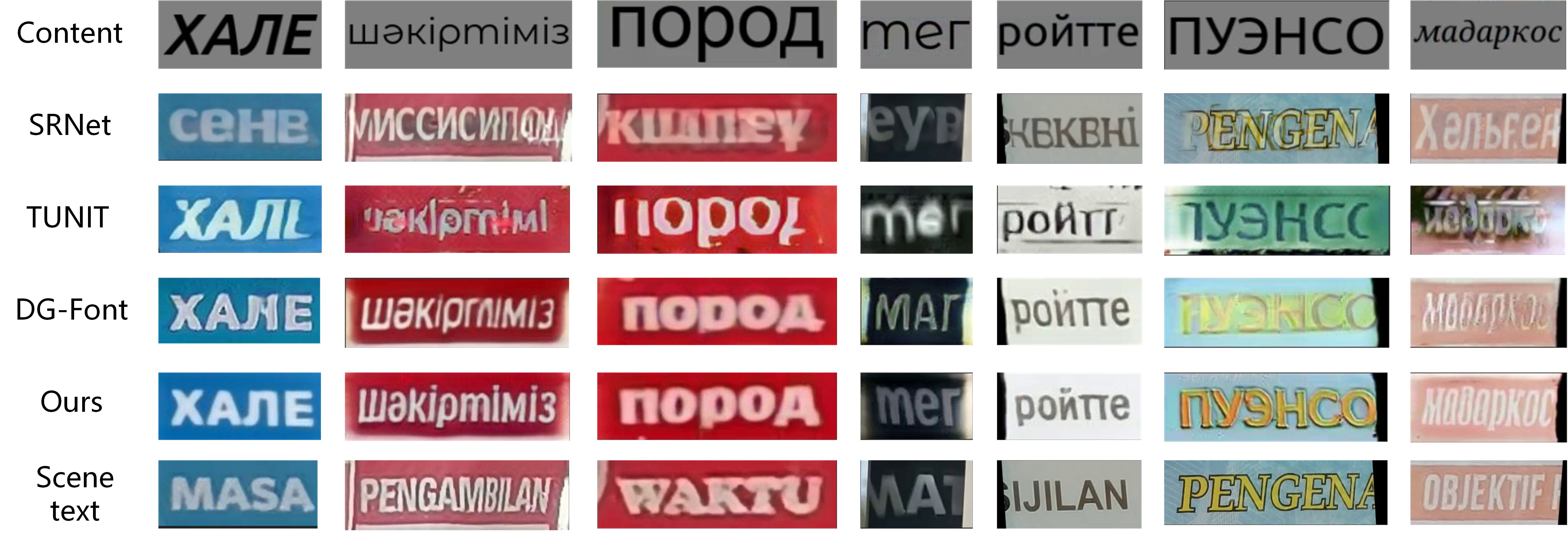}
\end{center}
   \caption{\textbf{Quantitative evaluation of Kazakh image generation.}}
   \label{fig4}
\end{figure*}

\begin{figure*}[htbp]
\label{qualititive_results}
\begin{center}
   \includegraphics[width=\linewidth]{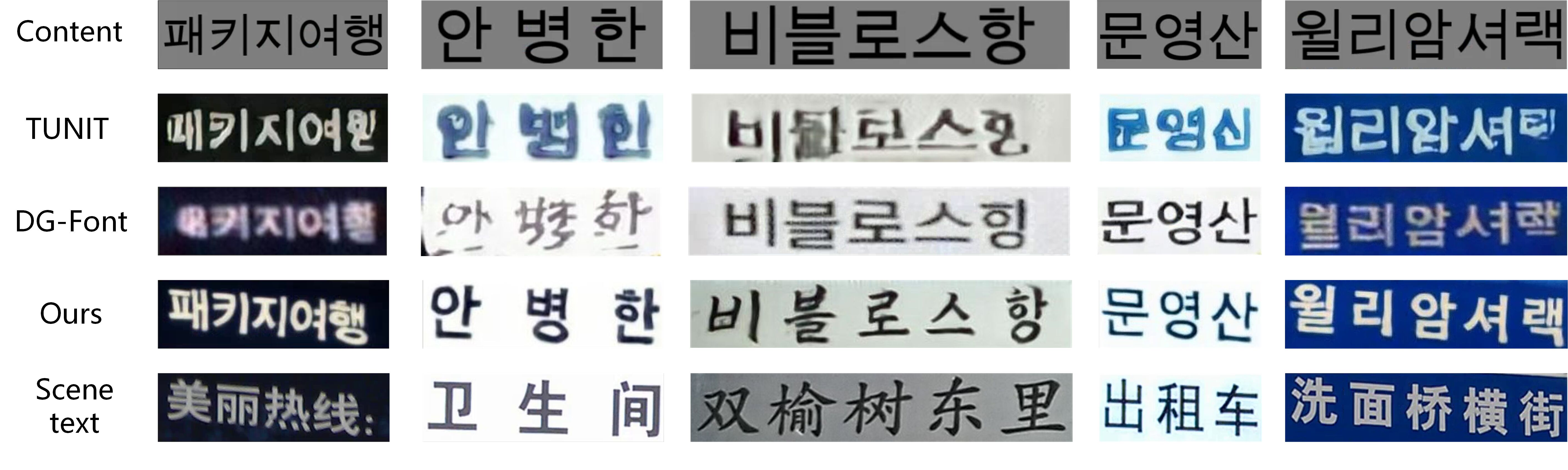}
\end{center}
   \caption{\textbf{Quantitative evaluation of Korean image generation.}}
   \label{fig5}
\end{figure*}

\subsection{Comparison with State-of-the-art Methods}
In this subsection, we first present the quantitative results based on the TRBA \cite{baek2019STRcomparisons} model to show that our method is effective in alleviating the shortage of data for low-resource languages. Then we demonstrate the robustness of our proposed method by presenting results obtained from three different scene recognition models. After that, qualitative results are shown to describe the disadvantages of comparison methods. Last but not least, we designed the ablation study to investigate the effect of different parts. We compare our model with the following methods. 
\begin{itemize}
\item SRNet \cite{DBLP:conf/mm/WuZLHLDB19}: SRNet employs a text conversion module, and a background inpainting module to generate rendered text and text-erased images respectively. SRNet relies on pre-designed font libraries to generate training datasets.
\item TUNIT \cite{DBLP:journals/corr/abs-2006-06500}: TUNIT is an unsupervised image-to-image translation model which separates the content and style of natural animal images and combines them with an adaptive instance normalization layer.
\item DGFont \cite{DBLP:conf/cvpr/XieCSL21}: DGFont is an unsupervised font generation method that introduces the feature deformable skip connection to deform and transform the character of one font to another.
\end{itemize}  
We also explore the data augmentation used in \cite{DBLP:journals/pami/WangXLLLYLS22} and claim that data augmentation alone cannot solve the recognition problem of low-resource languages. To further evaluate generalizability of our methods across various scenes, we employ three state-of-the-art scene recognition models. 
\begin{itemize}
\item TRBA\cite{baek2019STRcomparisons}: TRBA is a previous time-tested state-of-the-art model and it is widely used in evaluating synthetic dataset \cite{DBLP:journals/corr/abs-2003-10608, DBLP:conf/icdar/YimKCP21}. The method introduces a
unified four-stage framework that most existing scene text recognition
models fit into. We choose the TPS-ResNet-BiLSTM-Attn architecture in our experiments.
\item PARSeq \cite{DBLP:conf/eccv/BautistaA22}: ParSeq 
combins the features from a Permutated Language Modelling (PLM) multi-head attention model, with the encoded features from a Visual Transformer backbone \cite{DBLP:conf/iclr/DosovitskiyB0WZ21}. Noted that this method requires a large number of training samples due to the architecture of transformer.
\item SCATTER\cite{DBLP:conf/cvpr/LitmanATLMM20}: SCATTER train deep BiLSTM encoders by stacking BiLSTM blocks with intermediate supervision that improving the encoding of contextual dependencies. Moreover, SCATTER utilizes selective decoders to process the encoded features more efficiently.
\end{itemize}

\subsubsection{Quantitative comparison} 
The quantitative results are shown in Table \ref{quantatitive_results}. The quantitative experiments are divided into three parts by horizontal lines. In the first part, we compare the TRBA model with and without augmentation used in \cite{DBLP:journals/pami/WangXLLLYLS22}. For this experiment, the recognition models are trained only on real datasets from Kazakh and Korean and $aug$ denotes the data with augmentation used in \cite{DBLP:journals/pami/WangXLLLYLS22}. The model with data augmentation achieves higher accuracy and edit distance but the improvement is not obvious. In the second part, each method generates 1 $M$ images, which we use to train recognition models respectively. It can be seen that the proposed methods outperform all comparison methods. Specifically, SRNet can hardly generate useful scene text images for recognition datasets. TUNIT and DGFont show promising results in Kazakh but are still impractical in Korean, whose text is more complex than Kazakh. In the third part, we replace half of the original 1 $M$ images generated by each method with other scene text images generated by SynthText. DGFont and our proposed method achieve better results when cooperating with SynthText. Our methods improve the results of SynthText by more than 10 points in recognition accuracy. 

To assess the generalizability of our methods, we utilized three scene text recognition methods and reported eight results for each method based on four distinct datasets: a real dataset, a dataset generated by our method, a dataset generated by SynthText, and a combination of the SynthText dataset and our generated dataset. As shown in Table \ref{general}, the recognition accuracy for Korean was poor when the recognition model was only trained with real datasets. While our method of generating datasets effectively improved the performance of all three recognition models. Since the real datasets contain only few samples, we employed data augmentation when training PARSeq with the real dataset to enhance its performance and denote it using $*$.

\begin{table}[ht]
\caption{Ablation study.}
\begin{tabular}{lllllll}
\toprule
Methods & Kazakh(Acc) & Kazakh(ED) & Korean(Acc) & Korean(ED) \\
\midrule
Full model  & 71.319 & 0.873 & 64.837 & 0.806\\
w/o global attention3  & 69.044 & 0.861 & 63.100  &0.792\\
w/o local attention1  & 63.509 & 0.837 & 55.596  & 0.739\\
w/o typeface classifier  & 59.440 & 0.815 & 54.202 & 0.736 \\
\bottomrule
\end{tabular}
\label{table:ablation}
\end{table}

\begin{figure*}[t]
\begin{center}
   \includegraphics[width=\linewidth]{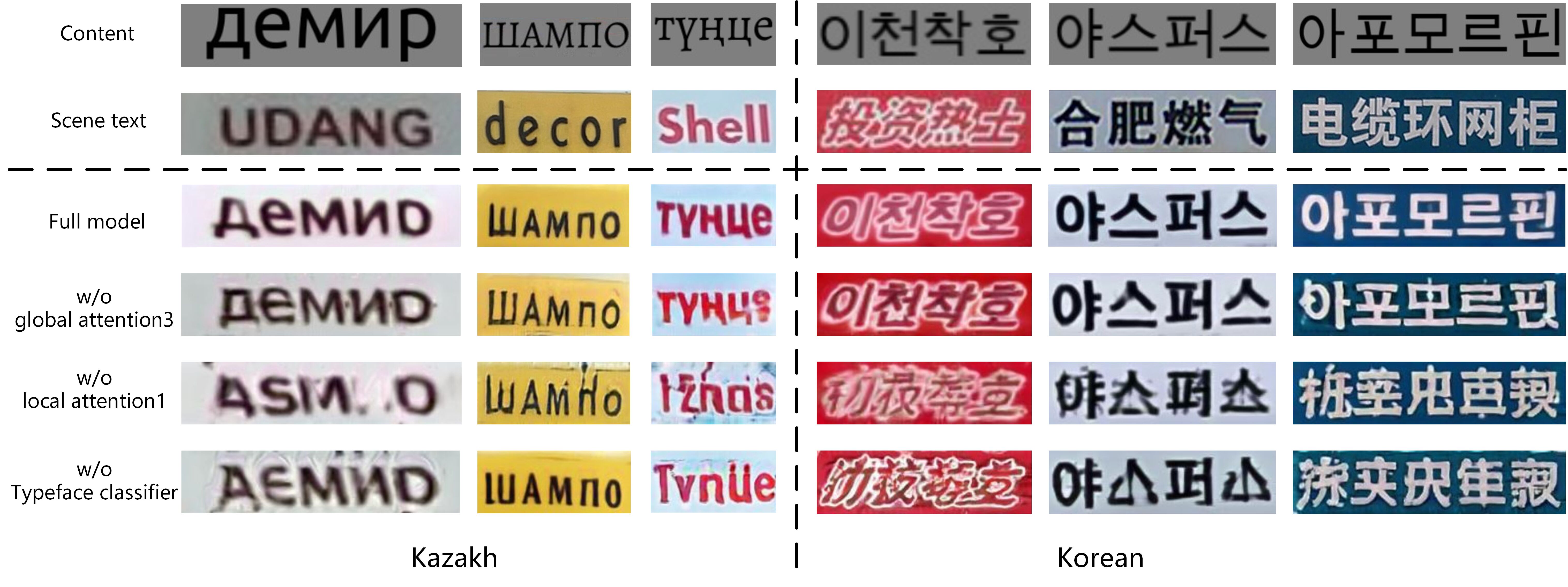}
\end{center}
   \caption{\textbf{Ablation study}.}
   \label{fig:ablation_images}
\end{figure*}

\subsubsection{Qualititive comparison}
In order to verify the capability of the proposed method to generate realistic scene text images, the visual comparison is shown in Figure \ref{fig4} and Figure \ref{fig5}. For each column of images, the first row shows the text content of generated images. The second rows to the penultimate row show the results generated by different methods. The last row shows the scene text image that provides the background and font styles. SRNet does not perform well in real-world images because it is trained on synthetic datasets. Besides, The images generated by SRNet are prone to blur due to the overlap between original content and new content. The images generated by TUNIT are often unreadable, and cannot be used in training the scene text recognition models. DGFont generates readable text images in easy cases, but it fails in complex cases which contain long text or diverse text styles. In contrast, the proposed method is able to generate realistic scene text images with clear content and similar font styles.

\subsection{Ablation study}
DGFont\cite{DBLP:conf/cvpr/XieCSL21}  and DGFont++\cite{DBLP:journals/corr/abs-2212-14742} have proven that inserting two deformable convolutions as skip connection helps maintain complete content of generate images. In this part, we remove different parts from the model successively and explore the influence, including the global attention3, local attention1, and typeface classifier. For each language, we conduct the ablation study on the data set of 1 $M$ synthetic images by our methods. Quantitative comparisons and quality comparisons are shown in Table \ref{table:ablation} and Figure \ref{fig:ablation_images}.

\textbf{(1)Effectiveness of global attention3.} We first remove the global attention3 from the full model. We can see that the quantitative results decrease especially in Kazakh. Figure \ref{fig:ablation_images} shows that without the global attention3, the generated images are prone to lose details, which is obvious in alphabetical languages like Kazakh whose scripts are simple. For Korean whose text is more complex than Kazakh, the decrease of the accuracy and edit distance are less.

\textbf{(2)Effectiveness of local attention1.} Local attention help learn local relationships between content images and generated images. When we further remove the local attention, the generated images suffer from blurring and artifacts around text strokes. For the recognition tasks, the performance of accuracy and edit distance drops significantly in both Kazakh and Korean.

\textbf{(3)Effectiveness of typeface classifier.} Typeface classifier is introduced to bridge the script style gap between different languages. Without the typeface classifier, the model cannot generate reliable images.

\section{Conclusion}
In this paper, We propose a weakly supervised scene text generation method. By incorporating high-resource languages into the training process, the method enables the cross-language generation and is able to generate a large number of scene texts with different backgrounds and font styles for low-resource languages. Our method disentangles the content and style features of scene text images and preserves the complete content structure of generated images using an integrated attention module. Moreover, we incorporate a pre-trained font classifier to bridge the style gap between different languages. We evaluate our method using state-of-the-art scene text recognition models and show that our generated scene text significantly improves scene text recognition accuracy, and can complement other methods to achieve even higher accuracy. Our method shows promise in alleviating the problem of collecting sufficient datasets for low-resource languages, as it requires only a limited number of recognition-level labels for low-resource languages.

\section{Acknowledgements}
This work was supported by the National Key Research and Development Program of China under Grant No. 2020AAA0107903. 







\printcredits

\bibliographystyle{plain}

\bibliography{cas-sc-template}

\begin{thebibliography}{10}

\bibitem{DBLP:conf/iccvw/Atienza21}
Rowel Atienza.
\newblock Data augmentation for scene text recognition.
\newblock In {\em {IEEE/CVF} International Conference on Computer Vision
  Workshops, {ICCVW} 2021, Montreal, BC, Canada, October 11-17, 2021}, pages
  1561--1570. {IEEE}, 2021.

\bibitem{baek2019STRcomparisons}
Jeonghun Baek, Geewook Kim, Junyeop Lee, Sungrae Park, Dongyoon Han, Sangdoo
  Yun, Seong~Joon Oh, and Hwalsuk Lee.
\newblock What is wrong with scene text recognition model comparisons? dataset
  and model analysis.
\newblock In {\em International Conference on Computer Vision (ICCV)}, 2019.

\bibitem{DBLP:journals/corr/abs-2006-06500}
Kyungjune Baek, Yunjey Choi, Youngjung Uh, Jaejun Yoo, and Hyunjung Shim.
\newblock Rethinking the truly unsupervised image-to-image translation.
\newblock {\em CoRR}, abs/2006.06500, 2020.

\bibitem{DBLP:conf/eccv/BautistaA22}
Darwin Bautista and Rowel Atienza.
\newblock Scene text recognition with permuted autoregressive sequence models.
\newblock In Shai Avidan, Gabriel~J. Brostow, Moustapha Ciss{\'{e}},
  Giovanni~Maria Farinella, and Tal Hassner, editors, {\em Computer Vision -
  {ECCV} 2022 - 17th European Conference, Tel Aviv, Israel, October 23-27,
  2022, Proceedings, Part {XXVIII}}, volume 13688 of {\em Lecture Notes in
  Computer Science}, pages 178--196. Springer, 2022.

\bibitem{DBLP:conf/iccv/ChanGZE19}
Caroline Chan, Shiry Ginosar, Tinghui Zhou, and Alexei~A. Efros.
\newblock Everybody dance now.
\newblock In {\em 2019 {IEEE/CVF} International Conference on Computer Vision,
  {ICCV} 2019, Seoul, Korea (South), October 27 - November 2, 2019}, pages
  5932--5941. {IEEE}, 2019.

\bibitem{DBLP:journals/csur/ChenJZLW21}
Xiaoxue Chen, Lianwen Jin, Yuanzhi Zhu, Canjie Luo, and Tianwei Wang.
\newblock Text recognition in the wild: {A} survey.
\newblock {\em {ACM} Comput. Surv.}, 54(2):42:1--42:35, 2021.

\bibitem{DBLP:journals/corr/abs-2212-14742}
Xinyuan Chen, Yangchen Xie, Li~Sun, and Yue Lu.
\newblock Dgfont++: Robust deformable generative networks for unsupervised font
  generation.
\newblock {\em CoRR}, abs/2212.14742, 2022.

\bibitem{DBLP:journals/tip/ChenXYT20}
Xinyuan Chen, Chang Xu, Xiaokang Yang, and Dacheng Tao.
\newblock Long-term video prediction via criticization and retrospection.
\newblock {\em {IEEE} Trans. Image Process.}, 29:7090--7103, 2020.

\bibitem{DBLP:conf/iclr/DosovitskiyB0WZ21}
Alexey Dosovitskiy, Lucas Beyer, Alexander Kolesnikov, Dirk Weissenborn,
  Xiaohua Zhai, Thomas Unterthiner, Mostafa Dehghani, Matthias Minderer, Georg
  Heigold, Sylvain Gelly, Jakob Uszkoreit, and Neil Houlsby.
\newblock An image is worth 16x16 words: Transformers for image recognition at
  scale.
\newblock In {\em 9th International Conference on Learning Representations,
  {ICLR} 2021, Virtual Event, Austria, May 3-7, 2021}. OpenReview.net, 2021.

\bibitem{DBLP:conf/icdar/GomezSGNVMBK17}
Raul Gomez, Baoguang Shi, Lluis Gomez{-}Bigorda, Lukas Neumann, Andreas Veit,
  Jiri Matas, Serge~J. Belongie, and Dimosthenis Karatzas.
\newblock {ICDAR2017} robust reading challenge on coco-text.
\newblock In {\em 14th {IAPR} International Conference on Document Analysis and
  Recognition, {ICDAR} 2017, Kyoto, Japan, November 9-15, 2017}, pages
  1435--1443. {IEEE}, 2017.

\bibitem{DBLP:conf/nips/GoodfellowPMXWOCB14}
Ian~J. Goodfellow, Jean Pouget{-}Abadie, Mehdi Mirza, Bing Xu, David
  Warde{-}Farley, Sherjil Ozair, Aaron~C. Courville, and Yoshua Bengio.
\newblock Generative adversarial nets.
\newblock In Zoubin Ghahramani, Max Welling, Corinna Cortes, Neil~D. Lawrence,
  and Kilian~Q. Weinberger, editors, {\em Advances in Neural Information
  Processing Systems 27: Annual Conference on Neural Information Processing
  Systems 2014, December 8-13 2014, Montreal, Quebec, Canada}, pages
  2672--2680, 2014.

\bibitem{DBLP:conf/cvpr/GuptaVZ16}
Ankush Gupta, Andrea Vedaldi, and Andrew Zisserman.
\newblock Synthetic data for text localisation in natural images.
\newblock In {\em 2016 {IEEE} Conference on Computer Vision and Pattern
  Recognition, {CVPR} 2016, Las Vegas, NV, USA, June 27-30, 2016}, pages
  2315--2324. {IEEE} Computer Society, 2016.

\bibitem{DBLP:conf/iccv/HeZRS15}
Kaiming He, Xiangyu Zhang, Shaoqing Ren, and Jian Sun.
\newblock Delving deep into rectifiers: Surpassing human-level performance on
  imagenet classification.
\newblock In {\em 2015 {IEEE} International Conference on Computer Vision,
  {ICCV} 2015, Santiago, Chile, December 7-13, 2015}, pages 1026--1034. {IEEE}
  Computer Society, 2015.

\bibitem{DBLP:conf/nips/HeuselRUNH17}
Martin Heusel, Hubert Ramsauer, Thomas Unterthiner, Bernhard Nessler, and Sepp
  Hochreiter.
\newblock Gans trained by a two time-scale update rule converge to a local nash
  equilibrium.
\newblock In {\em Advances in Neural Information Processing Systems 30: Annual
  Conference on Neural Information Processing Systems 2017, 4-9 December 2017,
  Long Beach, CA, {USA}}, pages 6626--6637, 2017.

\bibitem{DBLP:conf/iccv/HuangB17}
Xun Huang and Serge~J. Belongie.
\newblock Arbitrary style transfer in real-time with adaptive instance
  normalization.
\newblock In {\em {IEEE} International Conference on Computer Vision, {ICCV}
  2017, Venice, Italy, October 22-29, 2017}, pages 1510--1519. {IEEE} Computer
  Society, 2017.

\bibitem{DBLP:conf/cvpr/IsolaZZE17}
Phillip Isola, Jun{-}Yan Zhu, Tinghui Zhou, and Alexei~A. Efros.
\newblock Image-to-image translation with conditional adversarial networks.
\newblock In {\em 2017 {IEEE} Conference on Computer Vision and Pattern
  Recognition, {CVPR} 2017, Honolulu, HI, USA, July 21-26, 2017}, pages
  5967--5976. {IEEE} Computer Society, 2017.

\bibitem{DBLP:journals/corr/JaderbergSVZ14}
Max Jaderberg, Karen Simonyan, Andrea Vedaldi, and Andrew Zisserman.
\newblock Synthetic data and artificial neural networks for natural scene text
  recognition.
\newblock {\em CoRR}, abs/1406.2227, 2014.

\bibitem{DBLP:conf/eccv/JohnsonAF16}
Justin Johnson, Alexandre Alahi, and Li~Fei{-}Fei.
\newblock Perceptual losses for real-time style transfer and super-resolution.
\newblock In Bastian Leibe, Jiri Matas, Nicu Sebe, and Max Welling, editors,
  {\em Computer Vision - {ECCV} 2016 - 14th European Conference, Amsterdam, The
  Netherlands, October 11-14, 2016, Proceedings, Part {II}}, volume 9906 of
  {\em Lecture Notes in Computer Science}, pages 694--711. Springer, 2016.

\bibitem{DBLP:journals/corr/abs-2106-08385}
Praveen Krishnan, Rama Kovvuri, Guan Pang, Boris Vassilev, and Tal Hassner.
\newblock Textstylebrush: Transfer of text aesthetics from a single example.
\newblock {\em CoRR}, abs/2106.08385, 2021.

\bibitem{DBLP:journals/corr/abs-2107-11041}
Junyeop Lee, Yoonsik Kim, Seonghyeon Kim, Moonbin Yim, Seung Shin, Gayoung Lee,
  and Sungrae Park.
\newblock Rewritenet: Realistic scene text image generation via editing text in
  real-world image.
\newblock {\em CoRR}, abs/2107.11041, 2021.

\bibitem{DBLP:journals/chinaf/LiaoSLHYB20}
Minghui Liao, Boyu Song, Shangbang Long, Minghang He, Cong Yao, and Xiang Bai.
\newblock Synthtext3d: synthesizing scene text images from 3d virtual worlds.
\newblock {\em Sci. China Inf. Sci.}, 63(2):120105, 2020.

\bibitem{DBLP:conf/cvpr/LitmanATLMM20}
Ron Litman, Oron Anschel, Shahar Tsiper, Roee Litman, Shai Mazor, and
  R.~Manmatha.
\newblock {SCATTER:} selective context attentional scene text recognizer.
\newblock In {\em 2020 {IEEE/CVF} Conference on Computer Vision and Pattern
  Recognition, {CVPR} 2020, Seattle, WA, USA, June 13-19, 2020}, pages
  11959--11969. Computer Vision Foundation / {IEEE}, 2020.

\bibitem{DBLP:conf/iccv/0001HMKALK19}
Ming{-}Yu Liu, Xun Huang, Arun Mallya, Tero Karras, Timo Aila, Jaakko Lehtinen,
  and Jan Kautz.
\newblock Few-shot unsupervised image-to-image translation.
\newblock In {\em 2019 {IEEE/CVF} International Conference on Computer Vision,
  {ICCV} 2019, Seoul, Korea (South), October 27 - November 2, 2019}, pages
  10550--10559. {IEEE}, 2019.

\bibitem{DBLP:journals/corr/abs-2003-10608}
Shangbang Long and Cong Yao.
\newblock Unrealtext: Synthesizing realistic scene text images from the unreal
  world.
\newblock {\em CoRR}, abs/2003.10608, 2020.

\bibitem{DBLP:conf/icml/MeschederGN18}
Lars~M. Mescheder, Andreas Geiger, and Sebastian Nowozin.
\newblock Which training methods for gans do actually converge?
\newblock In Jennifer~G. Dy and Andreas Krause, editors, {\em Proceedings of
  the 35th International Conference on Machine Learning, {ICML} 2018,
  Stockholmsm{\"{a}}ssan, Stockholm, Sweden, July 10-15, 2018}, volume~80 of
  {\em Proceedings of Machine Learning Research}, pages 3478--3487. {PMLR},
  2018.

\bibitem{DBLP:journals/access/MuSXL21}
Deguo Mu, Wei Sun, Guoliang Xu, and Wei Li.
\newblock Random blur data augmentation for scene text recognition.
\newblock {\em {IEEE} Access}, 9:136636--136646, 2021.

\bibitem{DBLP:journals/eswa/NaiemiGK21}
Fatemeh Naiemi, Vahid Ghods, and Hassan Khalesi.
\newblock A novel pipeline framework for multi oriented scene text image
  detection and recognition.
\newblock {\em Expert Syst. Appl.}, 170:114549, 2021.

\bibitem{DBLP:conf/icdar/NayefLOPBCKKM0B19}
Nibal Nayef, Cheng{-}Lin Liu, Jean{-}Marc Ogier, Yash Patel, Michal Busta,
  Pinaki~Nath Chowdhury, Dimosthenis Karatzas, Wafa Khlif, Jiri Matas, Umapada
  Pal, and Jean{-}Christophe Burie.
\newblock {ICDAR2019} robust reading challenge on multi-lingual scene text
  detection and recognition - {RRC-MLT-2019}.
\newblock In {\em 2019 International Conference on Document Analysis and
  Recognition, {ICDAR} 2019, Sydney, Australia, September 20-25, 2019}, pages
  1582--1587. {IEEE}, 2019.

\bibitem{DBLP:conf/cvpr/NeumannM12}
Lukas Neumann and Jiri Matas.
\newblock Real-time scene text localization and recognition.
\newblock In {\em 2012 {IEEE} Conference on Computer Vision and Pattern
  Recognition, Providence, RI, USA, June 16-21, 2012}, pages 3538--3545. {IEEE}
  Computer Society, 2012.

\bibitem{DBLP:conf/cvpr/RoyBG020}
Prasun Roy, Saumik Bhattacharya, Subhankar Ghosh, and Umapada Pal.
\newblock {STEFANN:} scene text editor using font adaptive neural network.
\newblock In {\em 2020 {IEEE/CVF} Conference on Computer Vision and Pattern
  Recognition, {CVPR} 2020, Seattle, WA, USA, June 13-19, 2020}, pages
  13225--13234. Computer Vision Foundation / {IEEE}, 2020.

\bibitem{DBLP:conf/icdar/ShiYLYXCBLB17}
Baoguang Shi, Cong Yao, Minghui Liao, Mingkun Yang, Pei Xu, Linyan Cui,
  Serge~J. Belongie, Shijian Lu, and Xiang Bai.
\newblock {ICDAR2017} competition on reading chinese text in the wild
  {(RCTW-17)}.
\newblock In {\em 14th {IAPR} International Conference on Document Analysis and
  Recognition, {ICDAR} 2017, Kyoto, Japan, November 9-15, 2017}, pages
  1429--1434. {IEEE}, 2017.

\bibitem{DBLP:conf/nips/VaswaniSPUJGKP17}
Ashish Vaswani, Noam Shazeer, Niki Parmar, Jakob Uszkoreit, Llion Jones,
  Aidan~N. Gomez, Lukasz Kaiser, and Illia Polosukhin.
\newblock Attention is all you need.
\newblock In Isabelle Guyon, Ulrike von Luxburg, Samy Bengio, Hanna~M. Wallach,
  Rob Fergus, S.~V.~N. Vishwanathan, and Roman Garnett, editors, {\em Advances
  in Neural Information Processing Systems 30: Annual Conference on Neural
  Information Processing Systems 2017, December 4-9, 2017, Long Beach, CA,
  {USA}}, pages 5998--6008, 2017.

\bibitem{DBLP:journals/pami/WangXLLLYLS22}
Wenhai Wang, Enze Xie, Xiang Li, Xuebo Liu, Ding Liang, Zhibo Yang, Tong Lu,
  and Chunhua Shen.
\newblock {PAN++:} towards efficient and accurate end-to-end spotting of
  arbitrarily-shaped text.
\newblock {\em {IEEE} Trans. Pattern Anal. Mach. Intell.}, 44(9):5349--5367,
  2022.

\bibitem{DBLP:conf/mm/WuZLHLDB19}
Liang Wu, Chengquan Zhang, Jiaming Liu, Junyu Han, Jingtuo Liu, Errui Ding, and
  Xiang Bai.
\newblock Editing text in the wild.
\newblock In Laurent Amsaleg, Benoit Huet, Martha~A. Larson, Guillaume Gravier,
  Hayley Hung, Chong{-}Wah Ngo, and Wei~Tsang Ooi, editors, {\em Proceedings of
  the 27th {ACM} International Conference on Multimedia, {MM} 2019, Nice,
  France, October 21-25, 2019}, pages 1500--1508. {ACM}, 2019.

\bibitem{DBLP:journals/eswa/XiaoNSC22}
Zheng Xiao, Zhenyu Nie, Chao Song, and Anthony~Theodore Chronopoulos.
\newblock An extended attention mechanism for scene text recognition.
\newblock {\em Expert Syst. Appl.}, 203:117377, 2022.

\bibitem{DBLP:conf/cvpr/XieCSL21}
Yangchen Xie, Xinyuan Chen, Li~Sun, and Yue Lu.
\newblock Dg-font: Deformable generative networks for unsupervised font
  generation.
\newblock In {\em {IEEE} Conference on Computer Vision and Pattern Recognition,
  {CVPR} 2021, virtual, June 19-25, 2021}, pages 5130--5140. Computer Vision
  Foundation / {IEEE}, 2021.

\bibitem{DBLP:conf/cvpr/YangHL20}
Qiangpeng Yang, Jun Huang, and Wei Lin.
\newblock Swaptext: Image based texts transfer in scenes.
\newblock In {\em 2020 {IEEE/CVF} Conference on Computer Vision and Pattern
  Recognition, {CVPR} 2020, Seattle, WA, USA, June 13-19, 2020}, pages
  14688--14697. Computer Vision Foundation / {IEEE}, 2020.

\bibitem{DBLP:journals/tip/YaoBL14}
Cong Yao, Xiang Bai, and Wenyu Liu.
\newblock A unified framework for multioriented text detection and recognition.
\newblock {\em {IEEE} Trans. Image Process.}, 23(11):4737--4749, 2014.

\bibitem{DBLP:journals/pami/YeD15}
Qixiang Ye and David~S. Doermann.
\newblock Text detection and recognition in imagery: {A} survey.
\newblock {\em {IEEE} Trans. Pattern Anal. Mach. Intell.}, 37(7):1480--1500,
  2015.

\bibitem{DBLP:conf/icdar/YimKCP21}
Moonbin Yim, Yoonsik Kim, Hancheol Cho, and Sungrae Park.
\newblock Synthtiger: Synthetic text image generator towards better text
  recognition models.
\newblock In Josep Llad{\'{o}}s, Daniel Lopresti, and Seiichi Uchida, editors,
  {\em 16th International Conference on Document Analysis and Recognition,
  {ICDAR} 2021, Lausanne, Switzerland, September 5-10, 2021, Proceedings, Part
  {IV}}, volume 12824 of {\em Lecture Notes in Computer Science}, pages
  109--124. Springer, 2021.

\bibitem{DBLP:conf/cvpr/YinSL21}
Mingyu Yin, Li~Sun, and Qingli Li.
\newblock Id-unet: Iterative soft and hard deformation for view synthesis.
\newblock In {\em {IEEE} Conference on Computer Vision and Pattern Recognition,
  {CVPR} 2021, virtual, June 19-25, 2021}, pages 7220--7229. Computer Vision
  Foundation / {IEEE}, 2021.

\bibitem{DBLP:journals/corr/abs-1901-09193}
Fangneng Zhan, Hongyuan Zhu, and Shijian Lu.
\newblock Scene text synthesis for efficient and effective deep network
  training.
\newblock {\em CoRR}, abs/1901.09193, 2019.

\bibitem{DBLP:conf/cvpr/ZhanZL19}
Fangneng Zhan, Hongyuan Zhu, and Shijian Lu.
\newblock Spatial fusion {GAN} for image synthesis.
\newblock In {\em {IEEE} Conference on Computer Vision and Pattern Recognition,
  {CVPR} 2019, Long Beach, CA, USA, June 16-20, 2019}, pages 3653--3662.
  Computer Vision Foundation / {IEEE}, 2019.

\bibitem{DBLP:conf/icml/ZhangGMO19}
Han Zhang, Ian~J. Goodfellow, Dimitris~N. Metaxas, and Augustus Odena.
\newblock Self-attention generative adversarial networks.
\newblock In Kamalika Chaudhuri and Ruslan Salakhutdinov, editors, {\em
  Proceedings of the 36th International Conference on Machine Learning, {ICML}
  2019, 9-15 June 2019, Long Beach, California, {USA}}, volume~97 of {\em
  Proceedings of Machine Learning Research}, pages 7354--7363. {PMLR}, 2019.

\bibitem{DBLP:conf/eccv/ZhangD18}
Hang Zhang and Kristin~J. Dana.
\newblock Multi-style generative network for real-time transfer.
\newblock In Laura Leal{-}Taix{\'{e}} and Stefan Roth, editors, {\em Computer
  Vision - {ECCV} 2018 Workshops - Munich, Germany, September 8-14, 2018,
  Proceedings, Part {IV}}, volume 11132 of {\em Lecture Notes in Computer
  Science}, pages 349--365. Springer, 2018.

\bibitem{DBLP:journals/eswa/ZhongLSPL22}
Dajian Zhong, Shujing Lyu, Palaiahankote Shivakumara, Umapada Pal, and Yue Lu.
\newblock Text proposals with location-awareness-attention network for
  arbitrarily shaped scene text detection and recognition.
\newblock {\em Expert Syst. Appl.}, 205:117564, 2022.

\bibitem{DBLP:conf/cvpr/ZhouHMD18}
Peng Zhou, Xintong Han, Vlad~I. Morariu, and Larry~S. Davis.
\newblock Learning rich features for image manipulation detection.
\newblock In {\em 2018 {IEEE} Conference on Computer Vision and Pattern
  Recognition, {CVPR} 2018, Salt Lake City, UT, USA, June 18-22, 2018}, pages
  1053--1061. Computer Vision Foundation / {IEEE} Computer Society, 2018.

\bibitem{DBLP:conf/iccv/ZhuCZLD19}
Xizhou Zhu, Dazhi Cheng, Zheng Zhang, Stephen Lin, and Jifeng Dai.
\newblock An empirical study of spatial attention mechanisms in deep networks.
\newblock In {\em 2019 {IEEE/CVF} International Conference on Computer Vision,
  {ICCV} 2019, Seoul, Korea (South), October 27 - November 2, 2019}, pages
  6687--6696. {IEEE}, 2019.

\bibitem{DBLP:conf/cvpr/ZhuHLD19}
Xizhou Zhu, Han Hu, Stephen Lin, and Jifeng Dai.
\newblock Deformable convnets {V2:} more deformable, better results.
\newblock In {\em {IEEE} Conference on Computer Vision and Pattern Recognition,
  {CVPR} 2019, Long Beach, CA, USA, June 16-20, 2019}, pages 9308--9316.
  Computer Vision Foundation / {IEEE}, 2019.

\end{thebibliography}

\bio{}
\endbio

\end{document}